\documentclass[10pt,twocolumn,letterpaper]{article}

\usepackage{cvpr}              

\usepackage[pagebackref,breaklinks,colorlinks,allcolors=black]{hyperref}
\hypersetup{urlcolor=blue}
\usepackage{graphicx}
\usepackage{booktabs}
\usepackage{multirow}
\usepackage{amsmath,amssymb}
\usepackage{xcolor}
\usepackage{subcaption}
\usepackage{enumitem}

\usepackage{dblfloatfix} 
\usepackage{placeins}    
\usepackage{cuted}
\usepackage{capt-of}

\usepackage{array}
\usepackage{makecell}

\def\paperID{*****} 
\def\confName{CVPR}
\def\confYear{2026}

\title{CrashSight: A Phase-Aware, Infrastructure-Centric Video Benchmark for Traffic Crash Scene Understanding and Reasoning}

\author{
 Rui Gan$^{1}$ \quad Junyi Ma$^{1}$ \quad Pei Li$^{2}$\textsuperscript{*} \quad Xingyou Yang$^{1}$ \quad Kai Chen$^{3}$ \quad Sikai Chen$^{1}$ \quad Bin Ran$^{1}$\\
$^{1}$University of Wisconsin--Madison $^{2}$University of Wyoming
$^{3}$Columbia University\\}
\begin{document}
\maketitle
\renewcommand{\thefootnote}{\fnsymbol{footnote}}
\footnotetext[1]{Corresponding author.}
\begin{abstract}

Cooperative autonomous driving requires traffic scene understanding from both vehicle and infrastructure perspectives. While vision-language models (VLMs) show strong general reasoning capabilities, their performance in safety-critical traffic scenarios remains insufficiently evaluated due to the ego-vehicle focus of existing benchmarks. To bridge this gap, we present \textbf{CrashSight}, a large-scale vision-language benchmark for roadway crash understanding using real-world roadside camera data. The dataset contains 250 crash videos, annotated with 13K multiple-choice question-answer pairs organized under a two-tier taxonomy. Tier 1 evaluates the visual grounding of scene context and involved parties, while Tier 2 probes higher-level reasoning, including crash mechanics, causal attribution, temporal progression, and post-crash outcomes. We benchmark 8 state-of-the-art VLMs and show that, despite strong scene description capabilities, current models struggle with temporal and causal reasoning in safety-critical scenarios. We provide a detailed analysis of failure scenarios and discuss directions for improving VLM crash understanding. The benchmark provides a standardized evaluation framework for infrastructure-assisted perception in cooperative autonomous driving. The CrashSight benchmark, including the full dataset and code, is accessible at \url{https://mcgrche.github.io/crashsight/}.
\end{abstract}

\begin{figure*}[t]
    \centering
    \includegraphics[width=\linewidth]{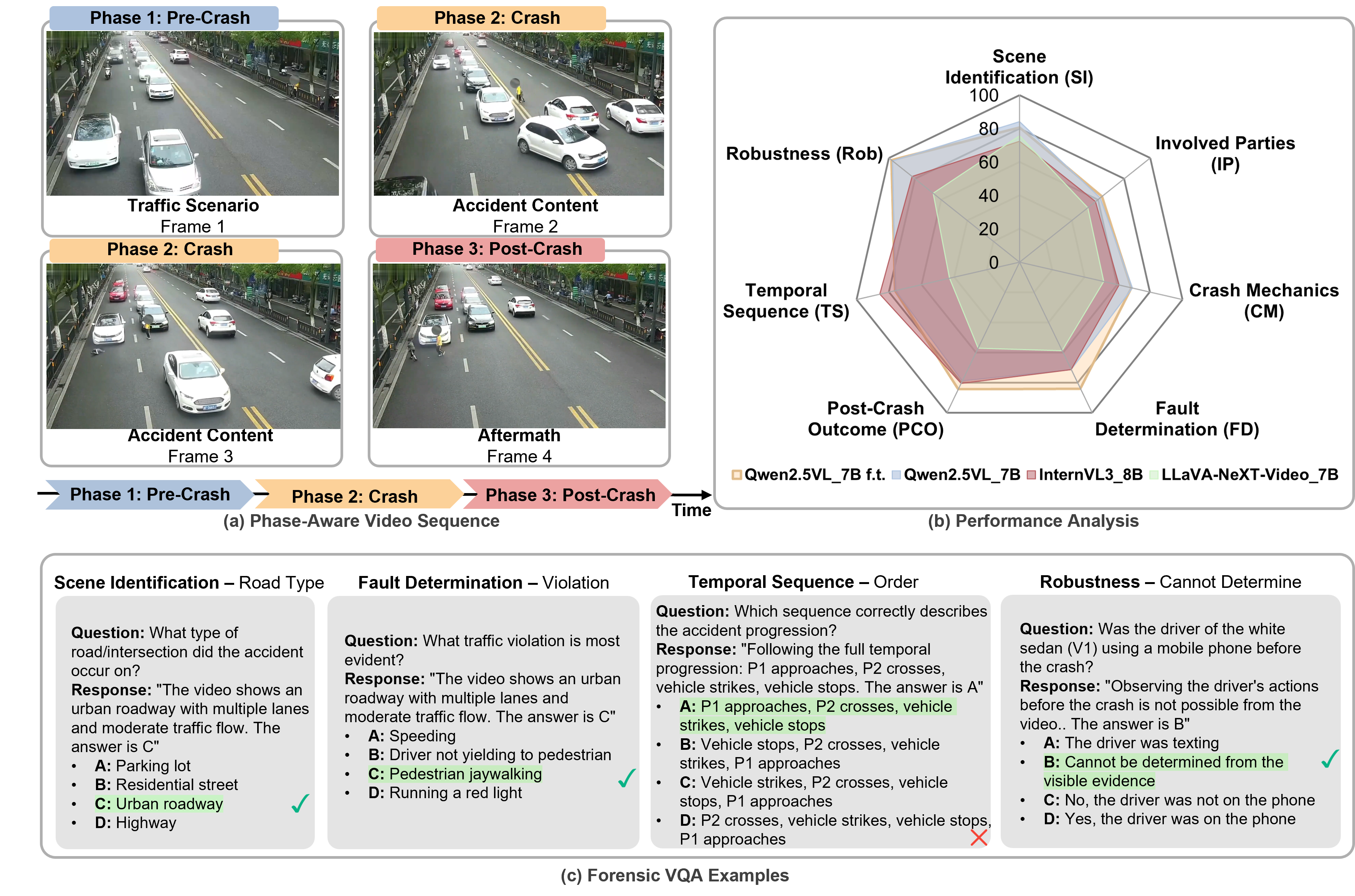}
    \caption{Overview of CrashSight-VQA. (a) Phase-aware temporal structure of a crash video. (b) VLM performance comparison across 7 QA categories (c) Example QA pairs spanning visual grounding and causal reasoning.}
    \label{fig:teaser}
\end{figure*}

\section{Introduction}
Cooperative autonomous driving (CDA) promises safer autonomous vehicles (AVs) by enabling them to share information with surrounding vehicles and infrastructure~\cite{you2026v2x}. Unlike single-vehicle autonomy that uses only onboard sensors, CDA integrates observations from both vehicle- and infrastructure-based sensors. This enables system-level situational awareness for safer AV decision-making.

CDA relies on models capable of understanding and explaining traffic scenes from both vehicle and infrastructure perspectives. Recently, foundation models, in particular, vision language models (VLMs), have emerged as powerful tools for multimodal traffic understanding and are increasingly explored for perception, explanation, and decision support in CDA. While prior studies demonstrate strong performance in general traffic understanding~\cite{wei2025driveqa,tian2025nuscenesspatialqa,zhou2025tumtrafficvideoqa}, VLMs' capabilities in safety-critical scenarios, i.e., crashes, remain largely unexplored. 

Crashes are critical long-tail events in traffic environments that demand a reliable understanding from any CDA system. Most existing studies adopt vehicle-centric viewpoints, advancing VLM-based crash reasoning using ego-vehicle cameras~\cite{xu2021sutd,kim2025vru,zhou2025tau106k}.
In contrast, VLM performance in understanding crashes from infrastructure-based sensors remains underrepresented. Current infrastructure-side studies are limited to anomaly detection without language supervision~\cite{sultani2018ucfcrime}, general traffic visual question answering (VQA) lacking crash scenarios~\cite{zhou2025tumtrafficvideoqa}, and large-scale surveillance benchmarks without transportation-specific semantics~\cite{liu2025surveillancevqa}. In addition, existing datasets lack crucial details for CDA tasks. These include structured temporal annotations or evaluation protocols targeting higher-level cognitive tasks such as causal reasoning, event progression understanding, and evidence-based safety assessment. To our knowledge, no prior benchmark integrates infrastructure-based crash videos with phase-aware temporal annotations and structured VQA tasks designed to evaluate VLMs in safety-critical traffic understanding.

To address this gap, we present \textbf{CrashSight}, a VQA benchmark that brings structured crash understanding to the infrastructure side.
The benchmark introduces a phase-aware annotation design that decomposes each crash into four temporal phases, pre-crash context, collision dynamics, aftermath, and potential causes, preserving the narrative structure that distinguishes crash events from routine traffic and enabling temporally grounded evaluation.
Building on expert-corrected dense captions, we construct 13K multiple-choice QA pairs spanning 7 categories: scene-level perception (scene identification, involved parties), event-level reasoning (temporal sequence, crash mechanics, post-crash outcome), and inference-level judgment (fault determination and robustness probes that test whether models can recognize when visual evidence is insufficient to support a conclusion).
We benchmark eight VLM configurations across four model families in both zero-shot and fine-tuned settings, revealing that domain-specific adaptation yields substantial accuracy gains while a persistent human-AI gap remains concentrated in visually demanding categories.
We further provide a systematic error taxonomy with transition analysis that traces persistent failures to architectural and training limitations, offering actionable directions for future model development.

The contributions of this work are fourfold: \textbf{1)} We introduce \textbf{CrashSight}, the first infrastructure-based crash VQA benchmark, comprising 250 expert-annotated surveillance clips with phase-aware dense captions and 13K multiple-choice QA pairs across seven categories.  \textbf{2)} We develop a 4-stage annotation pipeline combining VLM-assisted drafting, human expert refinement, and LLM-driven verification with augmentation, providing a scalable methodology for constructing benchmarks in safety-critical domains. \textbf{3)} We conduct comprehensive MLLM evaluation across eight configurations, demonstrating that fine-tuning yields up to +16.1 average accuracy improvement. \textbf{4)} We provide a systematic error taxonomy and transition analysis identifying visual token budget, frozen visual encoder, and pretraining distribution mismatch as the primary bottlenecks, with actionable implications for future benchmarks and model design.





\section{Related Work}\label{sec:related}

\noindent\textbf{Ego-view traffic scene understanding}
The dominant paradigm in traffic scene understanding adopts an ego-centric or dashcam viewpoint, driven primarily by autonomous-driving applications. Benchmarks and datasets such as DriveQA~\cite{wei2025driveqa}, NuScenes-SpatialQA~\cite{tian2025nuscenesspatialqa}, and MAPLM~\cite{cao2025maplm} evaluate general spatial reasoning and scene understanding from the vehicle's perspective. Differently, domain-adapted VLMs, including HazardVLM~\cite{xia2025hazardvlm}, CoT-VLM4Tar~\cite{ren2025cotvlm4tar}, and TrafficVLM~\cite{ding2025trafficvlm} extend these capabilities toward anomaly detection and hazard description.
When the focus shifts to crash-specific understanding, SUTD-TrafficQA~\cite{xu2021sutd} pioneered traffic video QA with 10K in-the-wild videos and 62.5K QA pairs spanning six reasoning tasks.
Subsequent benchmarks have substantially expanded both scale and scope:
VRU-Accident~\cite{kim2025vru} combines dense captioning with VQA for vulnerable road user (VRU) collision analysis;
RoadSocial~\cite{parikh2025roadsocial} curates diverse QA from social video narratives across 12 task types;
and TAU-106K~\cite{zhou2025tau106k} provides 106K clips for comprehensive crash understanding at scale.
On the model side, SafePLUG~\cite{sheng2025safeplug} introduces pixel-level grounding and temporal localization for crash analysis,
Fang~et~al.~\cite{fang2024abductive} formulate abductive reasoning for ego-view crash perception,
and InterAct-Video~\cite{vishal2025interact} targets reasoning-rich QA in urban traffic.
Despite this progress, the above benchmarks and models assume an ego-centric perspective that serves in-vehicle autonomy.
They do not address the infrastructure-side viewpoint required for V2X cooperative perception, post-incident analysis, or traffic management, where the camera is fixed and observes the scene from above or the side.

\noindent\textbf{Infrastructure-view traffic scene understanding}
Roadside cameras have a long history in event detection and understanding. Early datasets, including UCSD~Ped~\cite{li2013anomaly}, Avenue~\cite{lu2013avenue}, ShanghaiTech~\cite{liu2018shanghaitech}, and UCF-Crime~\cite{sultani2018ucfcrime}, have established large-scale benchmarks for abnormal event recognition, but none include textual annotations suitable for vision-language research. UCA~\cite{uca2024} addressed this gap by introducing approximately 20K textual descriptions for surveillance anomalies, yet it provides only narrative captions without an interactive QA evaluation framework. SurveillanceVQA-589K~\cite{liu2025surveillancevqa} represents the first large-scale surveillance VQA benchmark, offering 589K QA pairs across 12 cognitively diverse question types.
However, it targets general anomaly understanding (e.g., fighting, theft, vandalism) rather than traffic-related events. TUMTraffic-VideoQA~\cite{zhou2025tumtrafficvideoqa} is among the few benchmarks that use roadside cameras for spatio-temporal traffic understanding. However, this research focuses on understanding general traffic scenes rather than safety-critical events. The TAD corpus~\cite{xu2025tad} provides surveillance recordings of real-world collisions for detection purposes, yet it offers no QA or captioning annotations.
Table~\ref{tab:dataset_comparison} provides a detailed comparison with prior traffic crash datasets.


\begin{table}[t]
\centering
\setlength{\tabcolsep}{2.5pt}
\renewcommand\arraystretch{1.1}
\footnotesize
\newcommand{\cmark}{\checkmark}
\newcommand{\pmark}{$\circ$}
\begin{tabular}{@{}l c c r c c c c@{}}
\toprule
Dataset & Year & View & \#QA & Crash & VQA & DC & PA \\
\midrule
CTA~\cite{you2020traffic}                         & 2020 & Ego  & --     & \cmark & --     & --     & -- \\
DADA-2000~\cite{fang2019dada}                  & 2020 & Ego  & --     & \cmark & --     & --     & -- \\
SUTD-TQA~\cite{xu2021sutd}                    & 2021 & Mix   & 62.5K  & \cmark & \cmark & --     & -- \\
MM-AU~\cite{fang2024abductive}                      & 2024 & Ego  & 58.6K  & \cmark & \cmark & --     & -- \\
VRU-Accident~\cite{kim2025vru}                 & 2025 & Ego  & 6K     & \cmark & \cmark & \cmark & -- \\
TAU-106K~\cite{zhou2025tau106k}                & 2025 & Mix   & 106K   & \cmark & \cmark & --     & -- \\
RoadSocial~\cite{parikh2025roadsocial}         & 2025 & Mix   & 260K   & \pmark & \cmark & --     & -- \\
\midrule
CTAD~\cite{luo2023simulation}                        & 2023 & Infrastructure  & --     & \cmark & --     & --     & -- \\
TAD~\cite{xu2025tad}                           & 2025 & Infrastructure  & --     & \cmark & --     & --     & -- \\
TUMTraf-A~\cite{zhou2025tumtrafficvideoqa}     & 2025 & Infrastructure  & --     & \cmark & --     & --     & -- \\
TUMTraffic-VQA~\cite{zhou2025tumtrafficvideoqa} & 2025 & Infrastructure & 85K    & --     & \cmark & \cmark & -- \\
\midrule
\textbf{Ours} & \textbf{2025} & \textbf{Infrastructure} & \textbf{13K} & \textbf{\cmark} & \textbf{\cmark} & \textbf{\cmark} & \textbf{\cmark} \\
\bottomrule
\end{tabular}
\caption{Comparison with traffic crash video understanding datasets. \textbf{DC}: Dense Captioning,  \textbf{PA}: Phase-Aware temporal annotations.
\pmark\,=\,partially crash-focused.}
\label{tab:dataset_comparison}
\end{table}


\section{Benchmark Construction}\label{sec:benchmark}

CrashSight is designed to evaluate VLM performance on crash understanding and reasoning using roadside camera videos.
The benchmark comprises 250 expert-annotated surveillance video clips, each accompanied by a phase-aware dense caption and a set of multiple-choice QA pairs spanning seven categories.
A key design principle is that the temporal phase structure of each crash, including pre-crash context, collision dynamics, post-crash aftermath, and expert causal analysis, is preserved from dense captioning through QA generation and verification.

\subsection{Data Source and Curation}\label{sec:data_source}

\begin{figure*}[t]
    \centering
    \includegraphics[width=\linewidth]{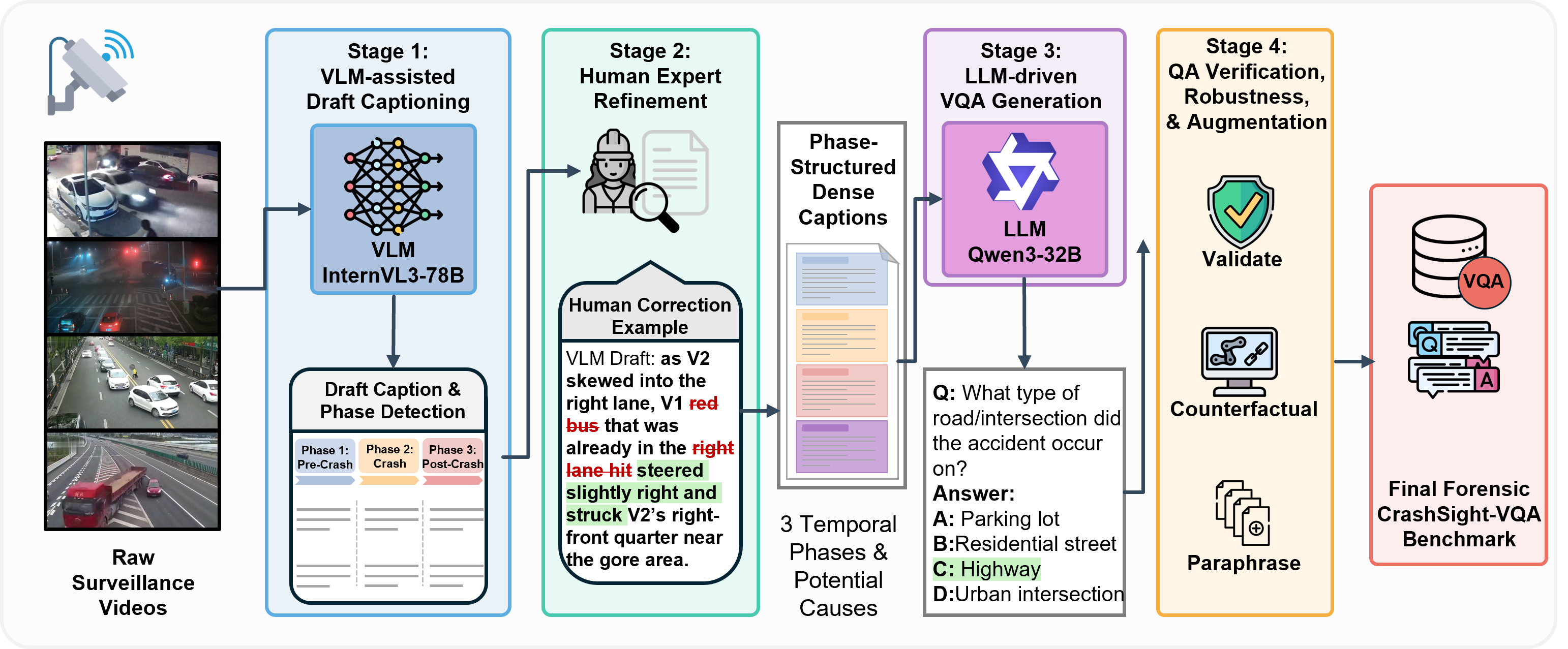}
    \caption{Overview of the CrashSight benchmark curation pipeline.
    Surveillance videos are processed through a three-stage annotation pipeline:
    (1)~VLM-assisted draft captioning with explicit phase boundaries,
    (2)~human expert refinement with standardized terminology, and
    (3)~LLM-driven VQA generation with counterfactual distractors, followed by verification and augmentation.
    Approximately 90\% of VLM drafts require substantial human correction.}
    \label{fig:pipeline}
\end{figure*}

Figure~\ref{fig:pipeline} illustrates the complete curation pipeline. We first source video clips from the TAD corpus~\cite{xu2025tad}, which contains real-world crash recordings captured by roadside cameras at various locations.
We select 250 clips that meet three criteria:
(i)~a visible collision or near-miss involving at least one road user,
(ii)~sufficient pre-crash context to enable causal reasoning, and
(iii)~observable post-crash aftermath.
Clip durations range from approximately 5 to 70 seconds, with resolution varying by camera installation.
The dataset is partitioned into 60/20/20 train/validation/test splits with clip-level disjointness to prevent information leakage.
All annotations and evaluations follow this fixed split.

\subsection{Phase-Aware Dense Captioning Pipeline}\label{sec:captioning_pipeline}

Crashes inherently follow a temporal narrative: what preceded the collision, what happened during impact, and what followed.
We encode this structure through a three-stage annotation pipeline that produces phase-aware dense captions for each clip.

\noindent\textbf{Stage~1: VLM-Assisted Draft Captioning.}
We prompt InternVL3-80B~\cite{internvl3} with structured templates that explicitly request four temporal phases:
\textit{[Traffic Scenario]} describing the pre-crash road environment and vehicle movements;
\textit{[Crash Content]} detailing the collision dynamics, including vehicle trajectories and impact configuration;
\textit{[Aftermath]} covering the post-crash scene state; and
\textit{[Potential Causes]} providing expert-level causal analysis.
The prompt emphasizes the roadside camera context and instructs the model to use specific entity identifiers (e.g., ``V1 white sedan,'' ``P1 motorcyclist'') and directional references relative to the camera view.
The model receives the full video and generates a draft caption with phase-delimited boundaries.
We deploy the model via vLLM~\cite{kwon2023vllm} for efficient batch processing across all 250 clips.

\noindent\textbf{Stage~2: Human Expert Refinement.}
To enhance caption quality, three trained annotators independently review each draft caption using a standardized, four-dimensional correction template:
(i)~\emph{entity precision}---replacing vague descriptions with specific vehicle types, colors, and identifiers;
(ii)~\emph{spatial relation accuracy}---correcting approach directions, lane positions, and relative positions that VLMs frequently hallucinate under oblique surveillance viewpoints;
(iii)~\emph{phase boundary accuracy}---adjusting the temporal delineation between pre-crash, crash, and post-crash segments; and
(iv)~\emph{causal specificity}---ensuring the \textit{[Potential Causes]} phase identifies concrete contributing factors rather than generic statements.
Nearly 90\% of the drafts require substantial corrections across at least two of these dimensions, underscoring the gap between current VLM capabilities and high-quality annotations.

\noindent\textbf{Stage~3: Quality Verification.}
A final verification pass ensures internal consistency between phase boundaries and the described events, standardizes terminology across all 250 clips, and flags any remaining ambiguities for adjudication.

\subsection{VQA Generation}\label{sec:vqa_category}

\begin{figure}[t]
    \centering
    \includegraphics[width=\linewidth]{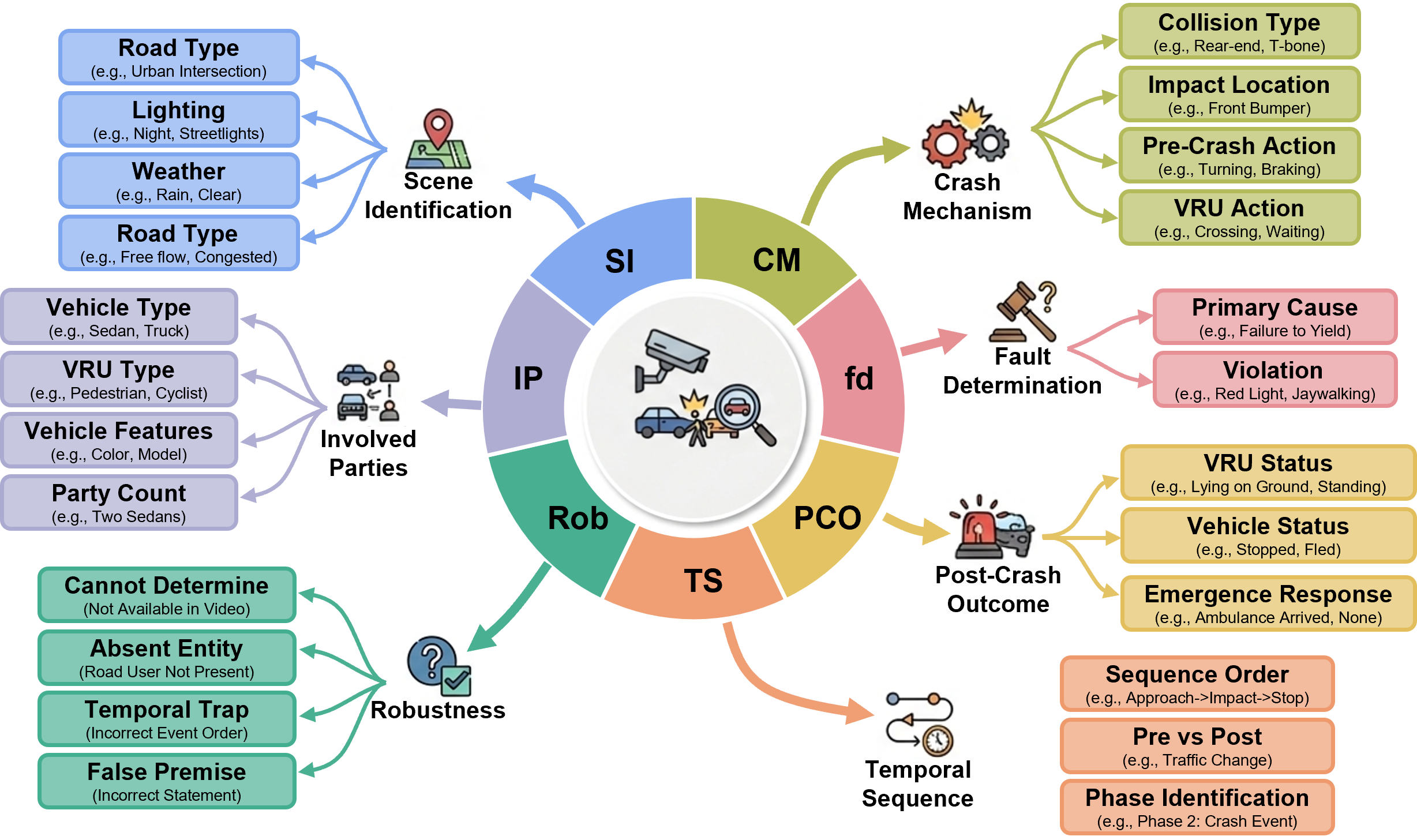}
    \caption{QA taxonomy of CrashSight. Seven categories are included: crash understanding requires phase-local recognition, while crash reasoning demands cross-phase temporal integration and causal inference. A robustness category probes hallucination resistance with four distinct question types.}
    \label{fig:qa_taxonomy}
\end{figure}

We design a two-stage QA generation pipeline that transforms phase-aware dense captions into multiple-choice questions across seven categories. These include six categories organized by cognitive demand and the temporal crash phases, with one robustness category that tests hallucination resistance (Figure~\ref{fig:qa_taxonomy}).

\smallskip
\noindent\textit{Tier~1---Crash Understanding} targets information recoverable from individual phases:

\begin{itemize}[nosep,leftmargin=1.2em]
\item \textbf{Scene Identification (SI):} road type, lighting, weather, and traffic context, answerable from Phase~1 (3--4 questions per video).
\item \textbf{Involved Parties (IP):} vehicle types,VRU types, identifying features, and party counts, requiring Phases~1 and~2 (3--4 questions).
\item \textbf{Post-Crash Outcome (PCO):} VRU status, vehicle condition, and emergency response indicators, answerable from Phase~3 (3--4 questions).
\end{itemize}

\noindent\textit{Tier~2---Crash Reasoning} demands cross-phase temporal integration and causal inference:

\begin{itemize}[nosep,leftmargin=1.2em]
\item \textbf{Crash Mechanics (CM):} collision type, impact dynamics, and pre-collision maneuvers, requiring detailed understanding of Phase~2 crash dynamics (3--4 questions).
\item \textbf{Fault Determination (FD):} primary cause and traffic violation identification, requiring synthesis across Phases~1, 2, and the expert causal analysis (2 questions).
\item \textbf{Temporal Sequence (TS):} event ordering, pre-vs-post comparisons, and phase identification, explicitly designed to be unanswerable from any single frame (2--3 questions).
\end{itemize}

\noindent\textit{Robustness} probes whether models can recognize the limits of observable evidence:

\begin{itemize}[nosep,leftmargin=1.2em]
\item Four hallucination probe types: \emph{cannot-determine} (e.g., driver BAC level, phone usage), \emph{absent-entity} (asking about road users not present), \emph{temporal-trap} (events outside the video timeframe), and \emph{false-premise} (questions with incorrect presuppositions).
The correct answer requires the model to resist fabricating unobservable information (5 questions per video).
\end{itemize}

Each question has one correct answer and three counterfactual distractors generated by an LLM conditioned on the question, correct answer, and category-specific distractor rules.
Correct answer positions are shuffled uniformly across A/B/C/D to mitigate position bias.

\noindent\textbf{Stage~1: Phase-grounded QA generation.}
For each video, we provide the four-phase caption to InternVL3-80B via vLLM with a structured prompt that specifies category templates, requires an \texttt{evidence} field grounded in the phase descriptions, and a \texttt{phase\_reference} field indicating which phase(s) support the answer.
This produces 14--18 raw QA pairs per video.

\noindent\textbf{Stage~2: Verification and augmentation.}
A second LLM pass performs three sub-tasks:
(i)~\emph{Verification}---each QA pair is checked for evidence accuracy, answer correctness, distractor plausibility, and phase reference consistency;
questions requiring information unobservable in surveillance footage (e.g., speed or velocity) are removed.
(ii)~\emph{Robustness augmentation}---five hallucination-probing questions are generated per video across the four probe types, with the ``cannot be determined'' option placed at randomized positions to prevent shortcut learning.
(iii)~\emph{Paraphrase augmentation}---each verified QA pair is paraphrased to increase linguistic diversity while preserving the correct answer and category label.
The final benchmark includes original, verified, robustness, and paraphrased variants.

\subsection{Benchmark Statistics}\label{sec:bench_stats}

\begin{figure}[t]
    \centering
    \includegraphics[width=\linewidth]{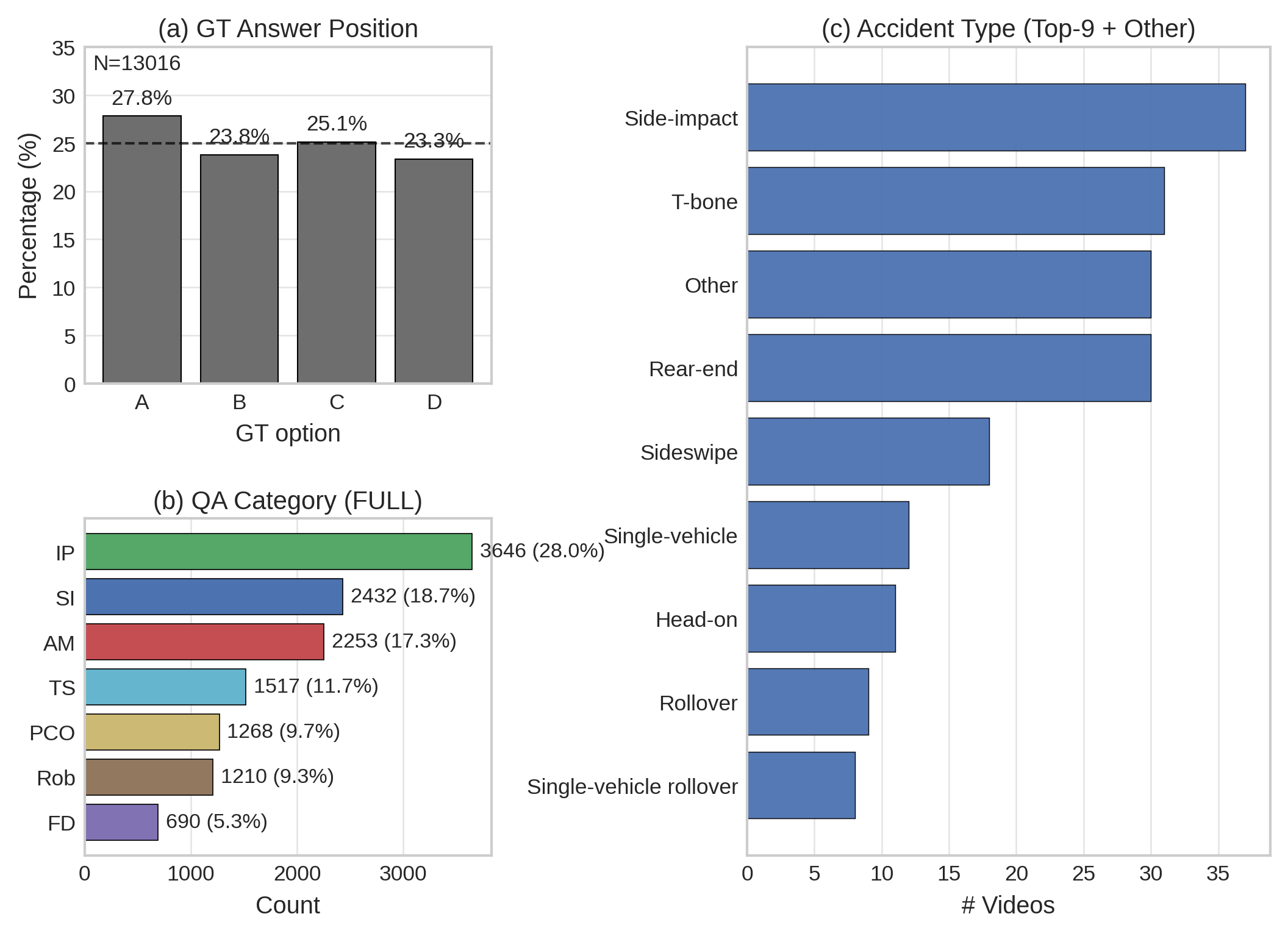}
    \caption{Dataset statistics of CrashSight-VQA.
    (a)~Ground-truth answer position distribution across all 13{,}016 QA pairs, showing approximate uniformity (23.3--27.8\%) after option shuffling.
    (b)~QA count by category.
    (c)~Distribution of accident types across the surveillance clips, with side-impact and T-bone collisions being most prevalent, reflecting the composition of the TAD source corpus.}
    \label{fig:dataset_stats}
\end{figure}

The complete CrashSight-VQA benchmark contains 250 annotated surveillance videos with expert-corrected phase-aware dense captions and approximately 13K multiple-choice QA pairs, totaling 4 candidate answer options.
Figure~\ref{fig:dataset_stats} reports per-category statistics grouped by cognitive tier.
Scene Understanding categories (SI, IP, PCO) exhibit relatively constrained answer vocabularies, reflecting the finite set of road types, vehicle types, and outcome states.
In contrast, Reasoning categories (CM, FD, TS) show substantially higher answer diversity and length, reflecting the open-ended nature of causal and temporal reasoning.
The Robustness subset contributes hallucination-probing questions across the four probe types described in Sec.~\ref{sec:vqa_category}.


\section{Experiments}

\begin{table*}[!t]
\centering
\caption{VQA accuracy (\%) on CrashSight-VQA. SI: Scene Identification, IP: Involved Parties, CM: Crash Mechanics, FD: Fault Determination, PCO: Post-Crash Outcome, TS: Temporal Sequence, Rob: Robustness. $\Delta$ is relative to our fine-tuned baseline. Best in \textbf{bold}, second-best \underline{underlined}.}
\label{tab:main_results}
\setlength{\tabcolsep}{2.6pt}
\renewcommand{\arraystretch}{1.05}
\begin{tabular}{l|c|c|ccccccc|c|c}
\toprule
Model & Size & Year & SI & IP & AM & FD & PCO & TS & Rob & AVG & $\Delta$ \\
\midrule
\multicolumn{12}{l}{\textit{Open-source Models}} \\
LLaVA-OneVision       & 0.5B & 2024 & 59.4 & 36.0 & 36.7 & 50.6 & 48.0 & 52.4 & 18.9 & 41.5 & -34.9 \\
LLaVA-NeXT-Video      & 7B   & 2024 & 75.5 & 52.3 & 51.6 & 58.6 & 57.1 & 42.9 & 66.1 & 58.6 & -17.7 \\
Qwen2.5VL         & 3B   & 2025 & 66.8 & 50.6 & 52.0 & 54.5 & 62.8 & 64.3 & 71.1 & 58.6 & -17.7 \\
Qwen2.5VL       & 7B   & 2025 & 67.7 & 51.9 & 55.4 & 66.7 & 74.2 & 66.7 & 73.3 & 62.9 & -13.5 \\
InternVL3      & 2B   & 2025 & 71.8 & 52.1 & 59.2 & 70.1 & 71.2 & \underline{81.0} & 71.1 & 64.2 & -12.1 \\
InternVL3       & 8B   & 2025 & 72.4 & 58.1 & 61.1 & 71.3 & \underline{80.3} & \textbf{85.7} & 82.2 & 68.7 & -7.7 \\
\midrule
\multicolumn{12}{l}{\textit{Fine-tuned Models (Ours)}} \\
Qwen2.5VL\_3B\_FT      & 3B   & 2025 & \textbf{84.0} & \underline{61.6} & \textbf{69.3} & \underline{71.3} & 78.8 & 76.0 & \underline{97.2} & \underline{74.7} & -1.6 \\
Qwen2.5VL\_7B\_FT      & 7B   & 2025 & \underline{80.6} & \textbf{63.2} & \underline{68.7} & \textbf{83.9} & \textbf{84.3} & 76.2 & \textbf{97.8} & \textbf{76.4} & +0.0 \\
\midrule
Human Expert          & --   & --   & 95.1 & 94.7 & 93.8 & 94.5 & 95.1 & 94.8 & 99.2 & 94.7 & +18.3 \\
\bottomrule
\end{tabular}
\end{table*}

\subsection{Experimental Setup}

We evaluate multiple-choice VQA task on the CrashSight benchmark.  
All reported results are computed on the held-out test split.

\noindent\textbf{Models.}
We benchmark a diverse set of VLMs in the zero-shot setting to characterize performance across model families and scales:
LLaVA-OneVision-0.5B~\cite{llava_ov}, Qwen2.5-VL-3B and 7B~\cite{qwen2vl}, LLaVA-NeXT-Video-7B~\cite{videomme2024}, and InternVL3-8B~\cite{internvl3}.
To measure the effect of domain-specific adaptation, we fine-tune both the 3B and 7B variants of Qwen2.5-VL-Instruct on our training split.
A human expert upper bound is established by three annotators who independently answer each test question, with the final answer determined by majority vote.

\noindent\textbf{Fine-tuning configuration.}
We adopt a QLoRA~\cite{dettmers2023qlora} protocol for fine-tuning VLMs.
The backbone is loaded in 4-bit NF4 quantization with \texttt{bfloat16} computation.
LoRA adapters with rank $r{=}16$, scaling factor $\alpha{=}32$, and dropout $0.05$ are injected into the query, key, and value projections as well as the MLP layers of the language backbone and the vision--language projector.
We use AdamW-8bit with a learning rate of $2{\times}10^{-4}$ and a cosine schedule with warmup ratio~$0.03$.
The per-device batch size is~1 with gradient accumulation over 8~steps (effective batch size~8).
Training proceeds for 2~epoch with a maximum sequence length of 8{,}192 tokens on a single NVIDIA A100 80\,GB GPU, requiring approximately 4~hours per model variant.
To encourage temporal awareness, we prepend a system prompt to every training sample that explicitly instructs the model to attend to how the scene \emph{changes across frames} and to ground its answer in observed video evidence.

\noindent\textbf{Video processing.}
We uniformly sample 4~frames per clip at 1\,FPS and resize within pixel bounds of $128{\times}28^{2}$ to $256{\times}28^{2}$ (${\approx}$100K--200K pixels per frame), consistent between training and evaluation.
All tasks share a unified instruction-following chat template in which the model receives the sampled frames and a textual prompt, and must generate either a structured phase-aware caption or a multiple-choice answer token (A/B/C/D).


\subsection{Experimental Results}\label{sec:vqa_results}

\begin{table*}[t]
\centering
\caption{Error taxonomy comparing vanilla Qwen2.5-VL (7B) vs fine-tuned Qwen2.5-VL (7B). Percentages are computed within each model's error set.}
\label{tab:error_taxonomy}
\resizebox{\textwidth}{!}{%
\begin{tabular}{l|rr|rr|r}
\toprule
Error Type & \multicolumn{2}{c|}{Vanilla (Qwen 7B)} & \multicolumn{2}{c|}{Fine-tuned (Qwen 7B ft.)} & $\Delta$ \\
\cmidrule(lr){2-3}\cmidrule(lr){4-5}
 & Count & \% & Count & \% & Count \\
\midrule
Unparseable output (no A/B/C/D) & 0 & 0.0 & 2 & 0.4 & 2 \\
Refusal / cannot determine (non-robustness) & 7 & 1.0 & 0 & 0.0 & -7 \\
Treats video as static image & 30 & 4.1 & 0 & 0.0 & -30 \\
Temporal confusion (TS errors) & 7 & 1.0 & 5 & 1.1 & -2 \\
Causal reasoning / fault attribution failure (FD errors) & 22 & 3.0 & 14 & 3.0 & -8 \\
Scene misidentification (SI errors) & 177 & 24.3 & 101 & 21.8 & -76 \\
Entity/party recognition error (IP errors) & 259 & 35.6 & 210 & 45.3 & -49 \\
Accident mechanics misunderstanding (CM errors) & 128 & 17.6 & 97 & 20.9 & -31 \\
Post-crash outcome misassessment (PCO errors) & 51 & 7.0 & 31 & 6.7 & -20 \\
Robustness failure (Rob errors) & 47 & 6.5 & 4 & 0.9 & -43 \\
\midrule
\textbf{TOTAL ERRORS} & 728 & 100.0 & 464 & 100.0 & -264 \\
\bottomrule
\end{tabular}%
}
\end{table*}

Table~\ref{tab:main_results} reports per-category accuracy across eight model configurations and a human expert upper bound.
All models are evaluated on the same held-out test split using identical multiple-choice prompts, deterministic decoding 
, and exact letter-match scoring.
We highlight four principal findings:

\noindent\textbf{Finding 1: Domain-specific fine-tuning yields substantial and consistent gains.}
Fine-tuning Qwen2.5-VL on our training split produces an average improvement of +16.1 points (3B) and +13.5 points (7B) over the corresponding vanilla models.
Gains are broad-based, spanning all seven categories, but are most significant for Robustness (+26.1 / +24.5), where fine-tuned models learn to select ``cannot be determined'' rather than hallucinating unobservable information, and for Crash Mechanics (+17.3 / +13.3), where exposure to phase-grounded collision descriptions improves crash-dynamics reasoning.
Notably, the fine-tuned 3B model (74.7\%) surpasses all zero-shot baselines, including InternVL3-8B (68.7\%), demonstrating that a small model with domain-specific adaptation can outperform a 4$\times$ larger general model on traffic understanding.

\noindent\textbf{Finding 2: Architecture matters more than scale for zero-shot reasoning.}
Among zero-shot baselines, InternVL3 consistently outperforms models of comparable or larger scale.
InternVL3-2B (64.2\%) surpasses Qwen2.5-VL-7B (62.9\%) despite being 3.5$\times$ smaller, and InternVL3-8B achieves the highest zero-shot average at 68.7\%.
The advantage is especially stark on Temporal Sequence, where InternVL3-8B attains 85.7\%, the highest score of \emph{any} model including fine-tuned variants. This suggests that its multi-image processing pipeline and training recipe offer strong temporal ordering capabilities even without domain-specific adaptation.

\noindent\textbf{Finding 3: Category difficulty reveals a reasoning hierarchy.}
We observe a clear difficulty gradient aligned with the cognitive tier structure defined in Sec.~\ref{sec:vqa_category}.
Involved Parties (IP) is the hardest category: even the best model achieves only 63.2\%, leaving a 31.5-point gap to human performance (94.7\%).
This difficulty stems from the need to jointly identify entity types, roles, and distinguishing features across multiple video frames under partial occlusion.
Crash Mechanics (CM) is similarly difficult (best: 69.3\%), as it requires understanding collision dynamics rather than static scene attributes.
In contrast, Post-Crash Outcome (PCO) and Fault Determination (FD) prove more accessible after fine-tuning (84.3\% and 83.9\% respectively), likely because aftermath states and causal labels recur across training examples with less visual ambiguity.

\subsection{Systematic Error Taxonomy}\label{sec:error_taxonomy}

To characterize the failure landscape of current VLMs on crash understanding, we classify incorrect results from both vanilla and fine-tuned Qwen2.5-VL-7B into 10 error types (Table~\ref{tab:error_taxonomy}).
A transition analysis across 1{,}962 test samples reveals that 1{,}117 (81.7\%) are answered correctly by both models, 381 (27.9\%) are newly corrected after fine-tuning, 117 (8.6\%) regress, and 347 (25.4\%) remain persistently incorrect.
Fine-tuning reduces total errors from 728 to 464 (36.3\%), with the clearest gains on failure modes tied to task format rather than visual understanding.
Three error types are completely or near-completely eliminated:
\emph{static image treatment}, where the vanilla model references ``the image shows'' instead of describing temporal dynamics; \emph{refusal on answerable questions}; and \emph{robustness hallucination}, as the fine-tuned model learns to select ``cannot be determined'' when evidence is genuinely absent.
Beyond these format-level corrections, fine-tuning also yields broad error reductions across categories, confirming that domain-specific adaptation improves both task comprehension and category-level reasoning.

However, the 347 persistently incorrect samples reveal structural limitations that fine-tuning cannot address.
These failures concentrate in Involved Parties (45.3\%), Scene Identification (21.8\%), and Accident Mechanics (20.9\%).
Fine-tuning further introduces 117 regressions (8.6\%), indicating that domain adaptation is not monotonically beneficial across all samples.

We attribute this resistance to three compounding factors rooted in the model architecture and training protocol.
\textbf{(i)~Insufficient visual token budget.}
The total visual information available to the model is bounded by frame count $\times$ per-frame pixel resolution.
Qwen2.5-VL processes only 8 uniformly sampled frames within $128$--$256 \times 28^{2}$ pixels per frame, yielding a fixed visual token budget regardless of video duration.
For video clips ranging from 5 to over 70 seconds, uniform sampling means that longer videos suffer progressively worse temporal coverage, and the key accident moment (often lasting fewer than 2 seconds) may fall entirely between sampled frames.
Simultaneously, the bounded pixel resolution compresses spatial detail in wide-angle overhead footage, making it difficult to resolve small, distant entities such as pedestrians, cyclists, or vehicle subtypes that are critical for Involved Parties and Accident Mechanics questions.
%
\textbf{(ii)~Quantization overhead.}
QLoRA's 4-bit NF4 quantization, while necessary for single-GPU fine-tuning, compresses visual encoder weight precision and may further degrade fine-grained feature discrimination for the subtle visual distinctions that entity recognition and mechanics understanding demand, such as differentiating a sedan from an SUV at 50 meters under oblique viewing geometry.
\textbf{(iii)~Frozen visual encoder.}
LoRA adapters are applied only to the language backbone and vision--language projector; the visual encoder itself remains frozen during fine-tuning.
The model therefore acquires improved \emph{verbal} reasoning about crash concepts (explaining why Robustness and Fault Determination improve markedly) but gains no new \emph{visual perception} capability for surveillance-specific challenges, including oblique camera angles, small distant objects, and partial occlusion.
%
Together, these factors explain why the residual error distribution concentrates so heavily on visually demanding categories, while format-level and knowledge-level errors are effectively resolved. Representative qualitative examples are available at \url{https://mcgrche.github.io/crashsight/}.


\section{Discussions}\label{sec:implications}


\textbf{Benchmark interpretation.} Although domain adaptation substantially improves overall performance, the remaining errors indicate that the dominant bottleneck is still visually grounded understanding under fixed-camera viewpoints, especially for questions that require precise identification of involved parties and reconstruction of crash mechanics. In this sense, the human expert baseline serves as a practical upper bound on the task and highlights that current models remain far from reliable performance, even when fine-tuned on domain data. At the same time, the benchmark should be interpreted with appropriate methodological care: the current annotation pipeline is effective for building a high-quality curated release, but it is not yet fully scalable because expert correction remains a major bottleneck in the loop. This limitation is especially important since the benchmark’s value depends heavily on the quality of its phase-aware captions and question design, meaning that larger-scale expansion cannot rely on automated generation alone without stronger verification and refinement mechanisms. 

Some categories, particularly Potential Causes and Fault Determination, are inherently more inferential than directly observable from surveillance video alone. Accordingly, labels in these categories are best understood as expert-validated judgments conditioned on the visible evidence in the clip, rather than as absolute legal truth. More broadly, the present release is still limited by dataset scale, source concentration, and the practical challenges of releasing surveillance-based crash data, all of which should be considered when interpreting benchmark performance and generalization claims. 


\textbf{Future directions.} Looking forward, the most immediate gains may come less from parameter scaling alone and more from targeted adaptation of model design, training strategy, and data utilization for domain-specific VLM settings, such as surveillance-based traffic crash reasoning~\cite{wei2025deepseek, cui2025paddleocr}. 

First, future research will focus on better visual token utilization, including adaptive frame sampling\cite{li2025improving}, event-driven frame selection around pre-crash and impact moments\cite{guo2025vtg}, and effective spatial resolution for decisive scenes. Moreover, a more deliberate co-design of training strategy and data regime is required \cite{ouyang2025conan}, since performance in this domain depends on how supervision, model adaptation, and data distribution interact with the unique perceptual and reasoning demands of crash videos. Beyond deciding how much data to use, future work should examine which mixtures of normal-traffic and accident data, as well as which optimization strategies, are best suited, including approaches that supervise not only final answer correctness but also the quality of evidence used, reasoning consistency, and structured explanation. 

Second, persistent failures in Involved Parties and Accident Mechanics suggest that domain adaptation alone is unlikely to be sufficient without more explicit spatial grounding, object tracking integration, and 3D-aware or motion-aware representations \cite{zheng2025video}. Beyond model design, benchmark evolution should also move from multiple-choice evaluation to richer task formats such as multi-turn question answering, dense phase-grounded captioning, and more complex questions that require structured explanations. 

Lastly, long-term progress of the \textbf{CrashSight} will require scaling both the number and diversity of videos. Future work will focus on collecting additional crash and normal-traffic videos using broader scene sources to reduce shortcut learning and improve robustness. Moreover, the accuracy and efficiency of the VQA generation pipeline can be further enhanced. This includes the usage of agentic workflows or high-performance VLMs, which may reduce substantial human efforts while preserving strong verification and refinement in the loop.


\section{Conclusion}\label{sec:conclusion}
While ego-view traffic understanding benchmarks have advanced rapidly, the infrastructure side has remained an open gap, hindering the advancement of CDA. 
We present \textbf{CrashSight}, an infrastructure-centric VQA benchmark designed to evaluate and enhance VLM performance on understanding traffic crash scenes. The dataset comprises 250 expert-annotated clips with over 13K multiple-choice QA pairs across seven categories. These include six categories focused on crash understanding and reasoning, and one robustness category that tests for hallucination resistance.
The dataset is produced using a three-stage annotation pipeline that combines VLM-assisted drafting, human expert refinement, and LLM-driven QA generation with verification.
A comprehensive evaluation of eight VLMs reveals that domain-specific fine-tuning yields an average accuracy improvement of at least +7.7\%.
Our systematic error taxonomy and transition analysis show that fine-tuning effectively resolves format-level and hallucination errors but leaves visually grounded perception failures partly untouched. Visual perception from roadside cameras, rather than language-level reasoning, emerges as the primary barrier for current foundation models.

\clearpage

{\small
\bibliographystyle{ieeenat_fullname}
\bibliography{main}
}

\clearpage

\end{document}